\documentclass[10pt,letterpaper]{paper}

\usepackage[all]{hypcap}

\usepackage[authoryear, round]{natbib}
\hypersetup{
    colorlinks = true,
    citecolor = {blue},
}

\usepackage{microtype}

\definecolor{good1}{RGB}{235,250,235}
\definecolor{good2}{RGB}{226,244,224}
\definecolor{good3}{RGB}{217,238,213}
\definecolor{good4}{RGB}{208,233,202}
\definecolor{good5}{RGB}{200,228,193}
\definecolor{good6}{RGB}{180,217,172}
\definecolor{good7}{RGB}{142,189,135}
\definecolor{good8}{RGB}{135,184,128}

\definecolor{bad1}{RGB}{252,243,240}
\definecolor{bad2}{RGB}{241,194,179}
\definecolor{tablerow}{HTML}{EDECE6}

\newcommand{\dgood}[2]{{\colorbox{good#1}{$#2$}}}
\newcommand{\dbad}[2]{{\colorbox{bad#1}{$#2$}}}
\newcommand{\dgoodbf}[2]{{\colorbox{good#1}{$\mathbf{#2}$}}}

\title{Tapered Language Models}

\author[1]{Reza Bayat}
\author[2]{Ali Behrouz}
\author[1,3,4]{Aaron Courville}

\affil[1]{Mila}
\affil[2]{Cornell University}
\affil[3]{Université de Montréal}
\affil[4]{CIFAR AI Chair}

\correspondingauthor{Reza Bayat at \email{reza.bayat@mila.quebec}.}

\begin{abstract}
\textbf{Abstract:} Modern language models, including transformer, recurrent, and memory-based variants, share a common chassis: a stack of identical layers in which parameters are allocated uniformly across depth.
This is a default inherited from the original transformer and largely unchanged since, yet a growing body of evidence suggests that layers contribute non-uniformly to the final output, with later layers refining the residual stream rather than transforming it.
We ask whether parameter capacity should reflect this asymmetry.
Our controlled experiment shows that, under a fixed budget, allocating more capacity to earlier layers and less to later layers improves perplexity over a uniform-width baseline, while the reverse allocation hurts.
Building on this result, we introduce \textit{Tapered Language Models} (TLMs), an architectural principle in which a parameter-bearing component is monotonically tapered across depth under a fixed total budget.
MLPs are the natural site for this instantiation: they dominate parameter count across all modern LM families and expose width as a single, clean axis of variation.
Across three model scales and four architectures (Transformer, Gated Attention, Hope-attention, and Titans), tapering MLP width via a smooth cosine schedule consistently improves perplexity and downstream benchmark performance over uniform baselines, at no additional parameter or compute cost.
These findings establish depth-aware capacity allocation as a simple, architecture-agnostic axis of language model design, a free lever hidden in plain sight.
\end{abstract}

\begin{document}

\maketitle

\section{Introduction}
\label{sec:intro}
\begin{figure}[!h]
    \centering
    \includegraphics[width=\linewidth]{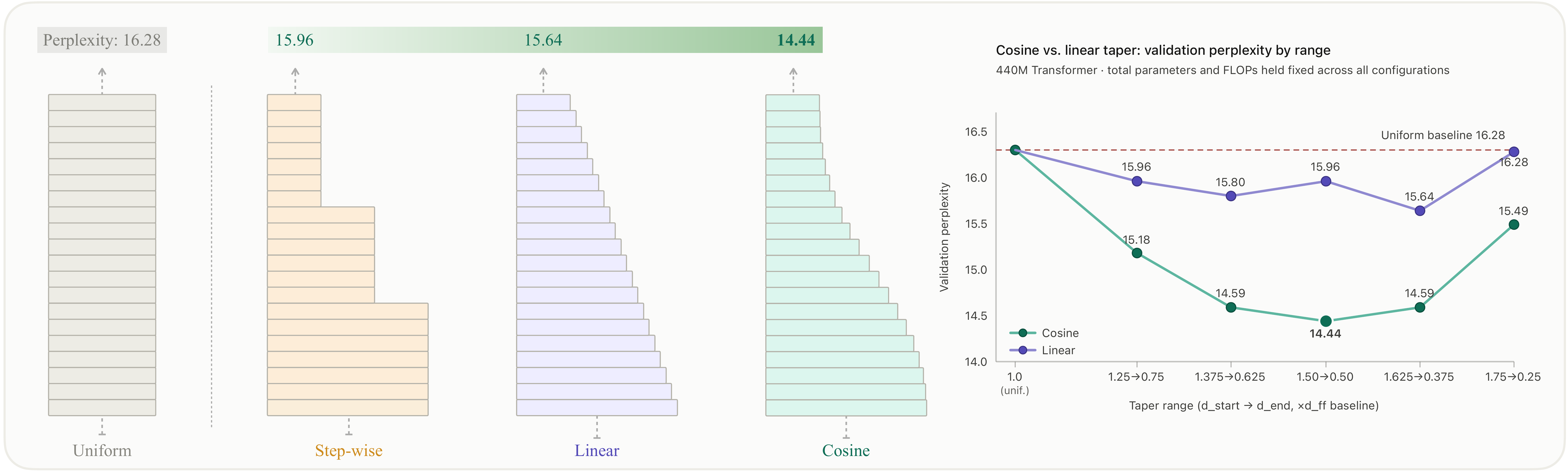}
    \caption{\textbf{Tapering MLP width improves perplexity at no additional parameter or compute cost.} \emph{Left:} per-layer MLP intermediate width for a uniform baseline and three tapered schedules (step-wise, linear, and cosine) on a $440$M Transformer; all configurations share the same total parameter count and FLOPs. In-distribution validation perplexity is shown above each profile. The cosine taper reaches $14.44$, improving over the uniform baseline ($16.28$) by $1.84$ points, with step-wise and linear in between. \emph{Right:} validation perplexity as a function of the taper range $d_{\text{start}}\!\rightarrow\!d_{\text{end}}$ (in units of $d_{\text{ff}}^{\text{baseline}}$) for the cosine and linear schedules. Both trace a U-shape that bottoms out at an intermediate ratio; cosine dominates linear at every range and is minimized at $1.50\!\rightarrow\!0.50$.}
    \label{fig:teaser}
\end{figure}

Modern language models, despite their surface diversity, share a common backbone. Transformers~\citep{vaswani2017attention}, gated-attention variants~\citep{qiu2025gated}, recurrent and state-space models~\citep{beck2024xlstm, orvieto2023resurrecting, peng2023rwkv, sun2023retentive, hasani2022liquid, dao2024transformers}, and memory-based architectures~\citep{schlag2021linear, behrouz2024titans, behrouz2025s, behrouz2025nested, behrouz2025atlas} differ in how they mix information across tokens, but they are all built from a stack of $L$ identical layers, each combining a token-mixing module with a feed-forward network (FFN). Within this backbone, parameters are distributed uniformly across depth: every layer receives the same allocation, regardless of its position in the stack. As models have scaled by orders of magnitude, this uniformity has remained largely unexamined, a default carried through from the original transformer~\citep{vaswani2017attention}.

There is, however, growing evidence that layers contribute non-uniformly to the final output. Early-exit methods show that the residual stream often converges to its final prediction well before the last layer~\citep{elbayad2020depth, belrose2023eliciting}, and layer-skipping frameworks demonstrate that later layers can be bypassed at inference time with minimal degradation in output quality~\citep{elhoushi2024layerskip}. Structured redundancy analyses find that many layers contribute negligibly to network function, and that deeper layers in particular can often be removed with surprisingly little performance loss~\citep{men2025shortgpt, gromov2024unreasonable, lad2024remarkable}. Interpretability work points in a similar direction: lower layers capture shallow syntactic patterns while upper layers encode more semantic ones~\citep{geva2021transformer}, suggesting that both the importance and the nature of computation shift across depth. Most of this evidence is drawn from transformers, but the pattern across these independent lines of work is consistent: layer importance is non-uniform.

\begin{figure}[t]
    \centering
    \includegraphics[width=\linewidth]{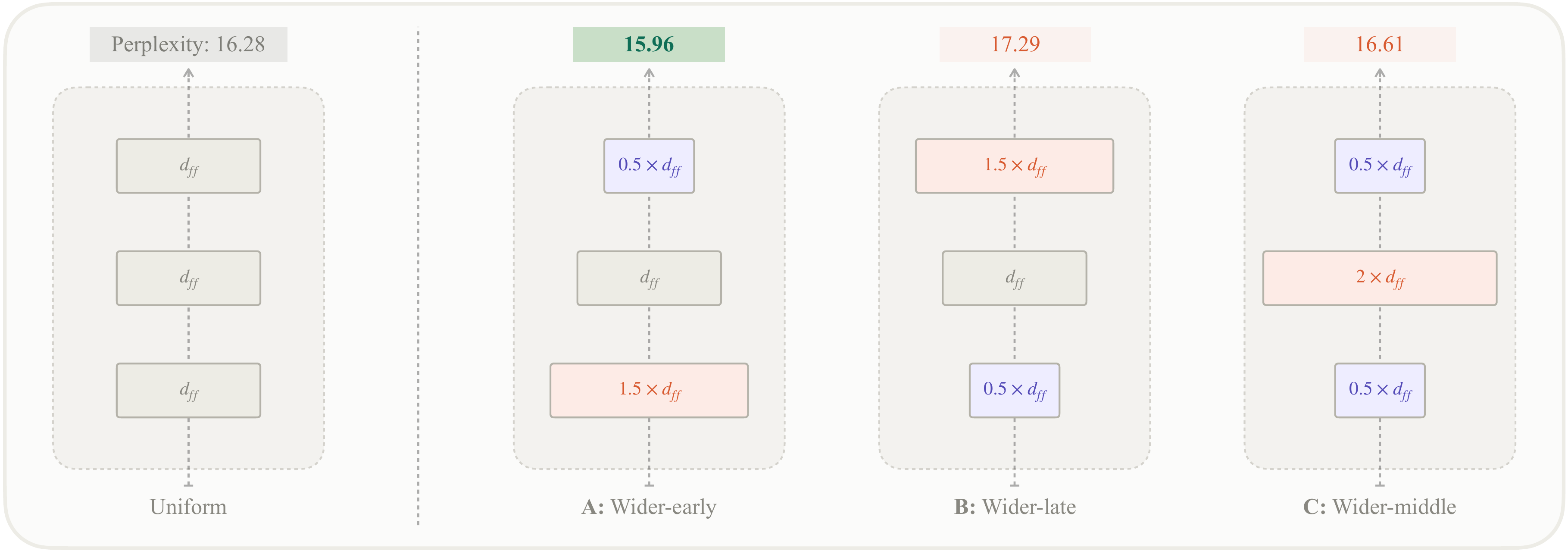}
    \caption{\textbf{Front-loading MLP capacity improves perplexity.} Layers of a 440M-parameter transformer are partitioned into three equal groups (early, middle, and late), and each group is assigned a different MLP intermediate width, with total parameter count held fixed across all configurations. Validation perplexity on a held-out split of the training data is reported above each configuration. Wider-early (A), which concentrates capacity in the early layers, achieves the lowest perplexity, improving over the uniform baseline by $0.32$ points. The reverse allocation, wider-late (B), substantially hurts performance, and concentrating capacity in the middle layers (C) also degrades it. Under the same parameter budget, the direction of capacity allocation changes perplexity by more than a point.}
    \label{fig:motivation}
\end{figure}

This raises a natural question: \textit{if layer importance is non-uniform across depth, why is layer capacity uniform?} To test whether the direction of capacity allocation matters, we conduct a simple experiment on a 440M-parameter transformer. We partition the model's layers into three equal groups and assign different MLP intermediate widths to each, while holding the total parameter count fixed across configurations. We compare four allocations: a uniform baseline; wider-early, in which capacity is concentrated in the early layers and reduced in later ones; wider-late, the reverse; and wider-middle, which concentrates capacity in the middle layers (Figure~\ref{fig:motivation}).

The result is clear. Wider-early achieves the lowest perplexity at $15.96$, improving over the uniform baseline by $0.32$ points. The reverse allocation, wider-late, hurts substantially, raising perplexity by over a full point to $17.29$, and concentrating capacity in the middle layers also degrades performance to $16.61$. The asymmetry is striking: under the same total parameter budget, the direction of allocation across depth changes perplexity by more than a point. Capacity is not a passive resource to be spread evenly across layers; it should be placed where it is needed most, which under this coarse three-block partition means front-loading it in the early layers.

The motivating experiment uses a coarse piecewise allocation with sharp transitions between groups, yet even this blunt reallocation confirms the value of front-loading capacity. A natural follow-up is whether smoother allocations do better, and how to specify them. We propose \textit{Tapered Language Models} (TLMs), an architectural principle in which a parameter-bearing component is monotonically tapered across depth under a fixed total budget, replacing sharp block transitions with a smooth decay from early to late layers. The principle is general: any depth-wise dimension that controls parameter count is a candidate, including attention head count, key-value dimension, recurrent state size, memory slots, and expert count in mixture-of-experts models~\citep{fedus2022switch}.

Among these candidates, MLPs are the most natural choice. They are the dominant parameter store in modern language models across architectural families, and their structure exposes a single width parameter, the intermediate dimension $d_{\text{ff}}$, that can be adjusted independently of the surrounding architecture. This choice is consistent with prior interpretability work characterizing FFNs as key-value memories whose contents shift from shallow to semantic patterns with depth~\citep{geva2021transformer}. We therefore taper $d_{\text{ff}}$ across layers using three smooth decay schedules (linear, cosine, and sigmoid), under a fixed total parameter budget.

While the motivating evidence for non-uniform layer importance is largely drawn from transformers, we find the prescription transfers. We evaluate TLMs on four architectures with substantially different token-mixing modules: standard Transformers~\citep{vaswani2017attention}, Gated Attention~\citep{qiu2025gated}, Hope-attention~\citep{behrouz2025nested}, and Titans~\citep{behrouz2024titans}. Across all four architectures and three scales (440M, 760M, and 1.3B parameters), tapering MLP width via the best of these schedules, a cosine decay, consistently improves perplexity and downstream benchmark performance over uniform-width baselines at matched parameters and FLOPs. That tapering helps in models whose token-mixing modules span softmax attention, gated attention, recurrent self-modifying memory, and neural long-term memory suggests the principle concerns how parameters are allocated across depth, rather than any property specific to attention. To probe why, we measure how much novel information each layer writes into the residual stream of pretrained transformers: MLP outputs become progressively more aligned with the existing residual as depth increases, reinforcing content already present rather than computing new features. Tapering aligns the architecture with this pattern by reducing capacity where it is least used.

Our contributions:
\begin{itemize}[leftmargin=2em, itemsep=0.5em]
    \item We introduce \textit{Tapered Language Models} (TLMs), an architectural principle in which a parameter-bearing component is monotonically tapered across depth under a fixed total budget; we instantiate this principle on MLP width with three smooth decay schedules (linear, cosine, and sigmoid) that preserve total parameter count and FLOPs.
    \item We demonstrate that tapering consistently improves perplexity and downstream benchmark performance across three model scales and four architecture families (Transformer, Gated Attention, Hope-attention, and Titans), establishing depth-aware capacity allocation as a robust, architecture-agnostic design choice.
    \item We provide a mechanistic analysis showing that MLP outputs become progressively more aligned with the residual stream at greater depths, directly motivating why front-loading capacity improves performance.
\end{itemize}

\section{Tapered Language Models}
\label{sec:method}

\subsection{Background}
\label{sec:background}

We focus on the family of decoder-only language models built from a stack of $L$ identical blocks, each composed of a token-mixing module $\mathcal{M}$ and a multilayer perceptron $\mathcal{F}$, both operating on a residual stream of dimension $d$. Given a hidden state $h_l \in \mathbb{R}^{N \times d}$ at layer $l$, the block computes
\begin{align}
z_l &= h_l + \mathcal{M}_l(h_l), \label{eq:attn} \\
h_{l+1} &= z_l + \mathcal{F}_l(z_l), \label{eq:ffn}
\end{align}
where layer normalization is omitted here for brevity. Architectures within this family differ in the choice of $\mathcal{M}$, ranging from softmax attention~\citep{vaswani2017attention} to gated attention~\citep{qiu2025gated} to recurrent and memory-based mechanisms~\citep{behrouz2024titans, behrouz2025nested}, but the MLP component $\mathcal{F}$ is structurally consistent across them: a depth-wise transformation parameterized by an intermediate dimension $d_{\text{ff}}$ that controls its capacity. Our analysis applies equally to gated variants such as SwiGLU~\citep{shazeer2020glu}, where $d_{\text{ff}}$ plays the same role despite the introduction of an additional projection; the per-layer MLP parameter count is $2\,d\,d_{\text{ff}}$ for vanilla FFNs and $3\,d\,d_{\text{ff}}$ for SwiGLU, both linear in $d_{\text{ff}}$. Typically, $d_{\text{ff}}$ is set to a fixed multiple of $d$ ($4d$ for standard FFNs or $\tfrac{8}{3}d$ for gated variants) and held constant across all $L$ layers.

\subsection{Tapering}
\label{sec:tapering}

The default choice of holding a per-layer architectural dimension constant across depth, while convenient, is one point in a larger design space. We define \textit{tapering} as the alternative in which a per-layer dimension is allowed to vary monotonically across depth under a fixed total budget.

Formally, consider any architectural component $C$ associated with a per-layer dimension $d_C(l)$ at layer $l \in \{0, 1, \ldots, L-1\}$. The uniform design fixes $d_C(l) = d_C^{\text{baseline}}$ for all $l$. A \textit{tapered} design instead requires
\begin{equation}
d_C(l+1) \leq d_C(l) \quad \text{for all } l, \qquad \frac{1}{L} \sum_{l=0}^{L-1} d_C(l) = d_C^{\text{baseline}},
\label{eq:tapering}
\end{equation}
so that capacity decreases (weakly) with depth while the average per-layer dimension, and therefore the total parameter budget associated with the component, is preserved.

This formulation is stated generally, applying to any depth-wise dimension that controls parameter count: attention head count, key-value dimension, recurrent state size, memory slot count, or expert count in mixture-of-experts models~\citep{fedus2022switch}. We instantiate the principle on MLP width in this work, and leave the empirical study of tapering along other dimensions to future work.

\subsection{Tapering MLP Width}
\label{sec:tapering-mlp}

The MLP component $\mathcal{F}$ in Equation~\ref{eq:ffn} is the dominant parameter store in modern language models, and its structure is shared across the architectures we consider, with $d_{\text{ff}}$ acting as a single width parameter that can be adjusted independently of the surrounding architecture. Tapering MLP width therefore yields a single, well-defined intervention that applies cleanly to transformers, gated attention models, and memory-based architectures alike.

\begin{figure}[t]
    \centering
    \includegraphics[width=\linewidth]{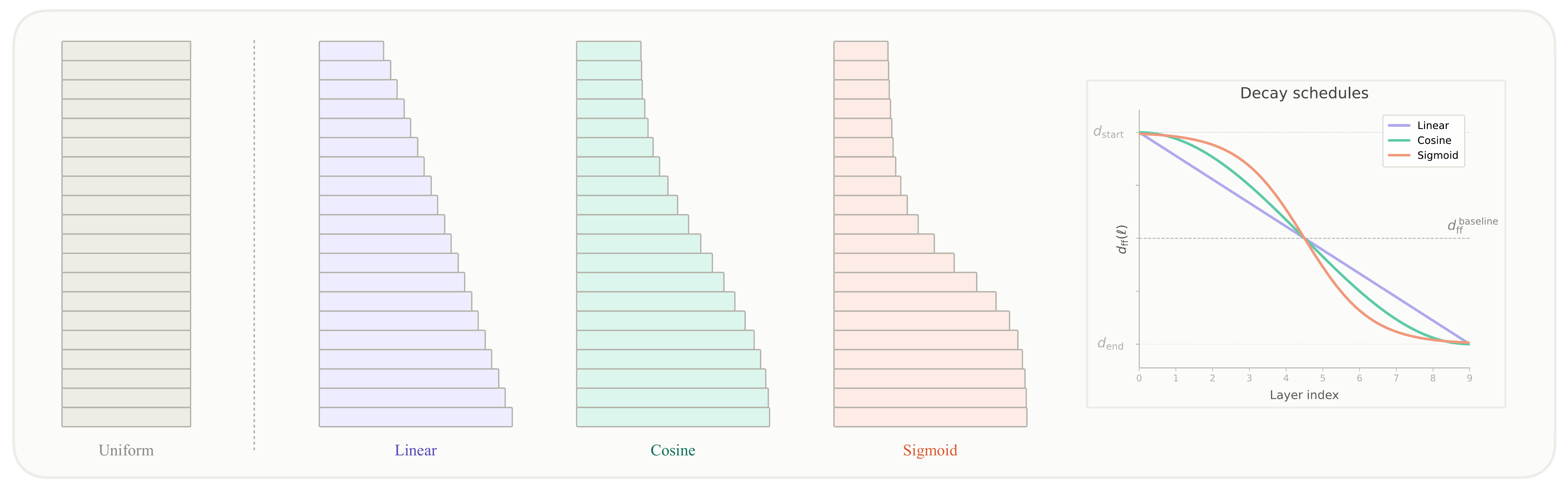}
    \caption{\textbf{Tapering smoothly reallocates MLP width across depth.} MLP intermediate dimension across layers for the uniform baseline and three tapered schedules, illustrated as per-layer bar widths (left) and as continuous decay curves (right) for a representative configuration. All configurations share the same total parameter count, and the curves cross the uniform baseline $d_{\text{ff}}^{\text{baseline}}$ at the midpoint since the per-layer widths average to $d_{\text{ff}}^{\text{baseline}}$ by construction.}
    \label{fig:schedules}
\end{figure}

Concretely, we replace the constant $d_{\text{ff}}$ with a per-layer intermediate dimension $d_{\text{ff}}(l)$ that decreases monotonically with layer index $l \in \{0, 1, \ldots, L-1\}$, parameterized by a start width $d_{\text{start}}$ and an end width $d_{\text{end}}$ with $d_{\text{start}} > d_{\text{end}}$. The piecewise allocation from the motivating experiment (Figure~\ref{fig:motivation}) follows the same principle but uses discrete block-wise assignment; we instead adopt smooth, continuous decay as a more natural and flexible generalization. We consider three decay schedules, illustrated in Figure~\ref{fig:schedules}.

\paragraph{Linear.} The intermediate dimension decreases at a constant rate:
\begin{equation}
\label{eq:linear}
d_{\text{ff}}(l) = d_{\text{start}} - (d_{\text{start}} - d_{\text{end}}) \cdot \frac{l}{L-1}.
\end{equation}

\paragraph{Cosine.} The dimension follows a half-cosine curve, decaying slowly near both endpoints and most steeply around the midpoint:
\begin{equation}
\label{eq:cosine}
d_{\text{ff}}(l) = d_{\text{end}} + \frac{d_{\text{start}} - d_{\text{end}}}{2} \left( 1 + \cos \frac{\pi l}{L-1} \right).
\end{equation}

\paragraph{Sigmoid.} The dimension follows a smooth step-down transition concentrated around a midpoint, with steepness controlled by $k > 0$ (we set $k = 10$ in all experiments):
\begin{equation}
\label{eq:sigmoid}
d_{\text{ff}}(l) = d_{\text{end}} + \frac{d_{\text{start}} - d_{\text{end}}}{1 + e^{k\left(\frac{l}{L-1} - 0.5\right)}}.
\end{equation}

The three schedules differ in how they distribute the transition across depth.
The linear schedule decays at a constant rate, engaging all layers equally but without
plateaus at either endpoint. The sigmoid schedule concentrates the change in a narrow band
around the midpoint, leaving most layers near $d_{\text{start}}$ or
$d_{\text{end}}$ and producing a near-binary allocation. The cosine schedule sits between
these extremes: its soft plateaus at both endpoints ensure a smooth entry and
exit, while its gradual mid-stack transition engages a wider range of
intermediate widths than either alternative. These geometric differences
directly predict the capacity profiles each schedule induces, and we return
to their empirical consequences in Section~\ref{sec:experiments}.

\paragraph{Parameter preservation.}
Specialized to MLP width, the budget-preservation condition in Equation~\ref{eq:tapering} requires the per-layer widths to average to the baseline:
\begin{equation}
\frac{1}{L} \sum_{l=0}^{L-1} d_{\text{ff}}(l) = d_{\text{ff}}^{\text{baseline}}.
\label{eq:preservation}
\end{equation}
We select $d_{\text{start}}$ and $d_{\text{end}}$ subject to this constraint, and the per-layer widths $d_{\text{ff}}(l)$ are then determined by the chosen schedule. The MLP at each layer involves projections between $d$ and $d_{\text{ff}}(l)$, so both its parameter count and its forward FLOP count are linear in $d_{\text{ff}}(l)$. As a result, the same averaging constraint that preserves total parameters also preserves total training and inference FLOPs: redistributing capacity across layers shifts where the FLOPs are spent but does not change the total.

\paragraph{Implementation details.}
The per-layer widths from each decay schedule are rounded to the nearest multiple of 16 to ensure efficient matrix operations on accelerator hardware. The first and last layer widths are fixed exactly to $d_{\text{start}}$ and $d_{\text{end}}$, and interior widths are adjusted in 16-unit increments to satisfy Equation~\ref{eq:preservation} exactly while preserving the monotonically decreasing order. Tapering is applied exclusively to the MLP intermediate dimension $d_{\text{ff}}$; other architectural hyperparameters (residual stream dimension $d$, attention head count, key-value dimension) remain identical to the uniform baseline.

\section{Experiments}
\label{sec:experiments}

\subsection{Setup}
\label{sec:setup}

\paragraph{Architectures.} We evaluate tapering on four architectures with substantially different token-mixing modules: Transformer~\citep{vaswani2017attention}, which uses standard softmax self-attention; Gated Attention~\citep{qiu2025gated}, which adds an output gating mechanism on top of softmax attention to remove attention sinks and improve sparsity; Hope-attention~\citep{behrouz2025nested}, a nested-learning architecture with self-modifying memory operating at multiple frequencies; and Titans~\citep{behrouz2024titans}, which augments attention with a neural long-term memory module that learns to memorize at test time.

\paragraph{Training.} We follow the standard pre-training setup of recent studies and use the Llama~3 tokenizer with a vocabulary size of $32$K and a training sequence length of $4$K tokens. The $440$M, $760$M, and $1.3$B models are trained on $30$B, $50$B, and $100$B tokens, respectively. We use AdamW with a cosine annealing schedule, a peak learning rate of $4 \times 10^{-4}$, a weight decay of $0.1$, and a global batch size of $0.5$M tokens. Training is performed on accelerator hardware. Optimizer states, learning rate, and all other training hyperparameters are held constant between uniform and tapered configurations; the only difference is the per-layer MLP intermediate dimension.

\paragraph{Evaluation.} We use two complementary perplexity setups across the experiments that follow. For schedule and width selection (§\ref{sec:sweep}), we report in-distribution validation perplexity on a held-out split of the training data. For the main results (§\ref{sec:main_results}), we report out-of-distribution perplexity on WikiText~\citep{merity2016pointer} and LAMBADA~\citep{paperno2016lambada}, alongside downstream accuracy on eight commonsense reasoning benchmarks: LAMBADA (accuracy), PIQA~\citep{bisk2020piqa}, HellaSwag~\citep{zellers2019hellaswag}, WinoGrande~\citep{sakaguchi2021winogrande}, ARC-easy and ARC-challenge~\citep{clark2018think}, SIQA~\citep{sap2019social}, and BoolQ~\citep{clark2019boolq}.

\subsection{Schedule and Width Sweep}
\label{sec:sweep}

We use the $440$M-parameter Transformer to identify a tapering configuration that we then carry forward, unchanged, to the main results. Two design choices govern the configuration: the decay schedule (linear, cosine, and sigmoid; Equations~\ref{eq:linear}--\ref{eq:sigmoid}) and the width ratio $d_{\text{start}}/d_{\text{end}}$, which controls how aggressively capacity is redistributed across depth. We sweep five width ratios, $\{1.25/0.75,\ 1.375/0.625,\ 1.5/0.5,\ 1.625/0.375,\ 1.75/0.25\}$, in combination with all three schedules, holding total parameter count fixed across all $15$ configurations as well as against the uniform baseline. Validation perplexity for every cell of this $5 \times 3$ sweep is reported in Table~\ref{tab:sweep}.

\begin{table}[t]
\centering
\setlength{\tabcolsep}{8pt}
\renewcommand{\arraystretch}{1.25}
\caption{\textbf{Schedule and width sweep.} In-distribution validation perplexity on the $440$M Transformer across five width ratios and three schedules. The uniform baseline achieves $16.28$ perplexity. The $\Delta$ columns report the change relative to the uniform baseline; green indicates improvement, red indicates regression, and saturation indicates magnitude. Total parameter count, training FLOPs, and inference FLOPs are held fixed across all $15$ configurations and against the uniform baseline.}
\label{tab:sweep}
\begin{tabular}{l cc cc cc}
\toprule
& \multicolumn{2}{c}{Cosine} & \multicolumn{2}{c}{Linear} & \multicolumn{2}{c}{Sigmoid} \\
\cmidrule(lr){2-3} \cmidrule(lr){4-5} \cmidrule(lr){6-7}
Taper range ($\times d_{\text{ff}}^{\text{baseline}}$) & ppl $\downarrow$ & $\Delta$ & ppl $\downarrow$ & $\Delta$ & ppl $\downarrow$ & $\Delta$ \\
\midrule
\addlinespace[2pt]
$1.25 \rightarrow 0.75$    & $15.18$           & \dgood{6}{-1.10} & $15.96$ & \dgood{2}{-0.32} & $16.44$ & \dbad{1}{+0.16} \\
$1.375 \rightarrow 0.625$  & $14.59$           & \dgood{7}{-1.69} & $15.80$ & \dgood{3}{-0.48} & $16.44$ & \dbad{1}{+0.16} \\
$1.50 \rightarrow 0.50$    & $\mathbf{14.44}$  & \dgoodbf{8}{-1.84} & $15.96$ & \dgood{2}{-0.32} & $16.12$ & \dgood{1}{-0.16} \\
$1.625 \rightarrow 0.375$  & $14.59$           & \dgood{7}{-1.69} & $15.64$ & \dgood{4}{-0.64} & $15.96$ & \dgood{2}{-0.32} \\
$1.75 \rightarrow 0.25$    & $15.49$           & \dgood{5}{-0.79} & $16.28$ & $\phantom{+}0.00$ & $17.12$ & \dbad{2}{+0.84} \\
\bottomrule
\end{tabular}
\end{table}

The empirical ordering is consistent with the geometric properties of the schedules described in Section~\ref{sec:method}. Across all five width ratios, cosine yields the lowest perplexity, linear the second-lowest, and sigmoid the worst---a strict ordering that holds even as the width ratio varies. The gap between cosine and the other two dominates the table: even cosine's worst configuration ($1.75/0.25$, $15.49$) outperforms linear's best ($1.625/0.375$, $15.64$). The ranking reflects how broadly each schedule engages the stack: linear decays at a constant rate with no endpoint plateaus; sigmoid concentrates its transition in a narrow midpoint band, effectively pinning most layers near $d_{\text{start}}$ or $d_{\text{end}}$; cosine occupies the middle ground, with soft endpoint plateaus and a gradual transition that puts a wider range of intermediate widths to use. That the middle option wins suggests the right allocation is neither uniform-rate nor near-binary, but a smooth gradient across depth.

Holding the schedule fixed at cosine, we then read off the optimal width ratio. Perplexity follows a clean U-shape across the five ratios, reaching its minimum at $1.5/0.5$ ($14.44$) and degrading on either side. The wider ratios ($1.625/0.375$ and $1.75/0.25$) push too much capacity into the early layers and starve the late layers; the narrower ratios ($1.375/0.625$ and $1.25/0.75$) leave the redistribution too mild to fully exploit the asymmetry. We adopt cosine with $d_{\text{start}}/d_{\text{end}} = 1.5/0.5$ for all subsequent experiments, fixed across architectures and scales. This sweep is conducted at a single operating point, and the U-shape minimum may sit elsewhere at other scales or architectures; the chosen configuration is therefore best read as a robust default rather than a global optimum, and the gains we report should be read as a lower bound on what the principle can deliver.

\subsection{Language Modeling and Commonsense Reasoning}
\label{sec:main_results}

We carry the cosine schedule with $d_{\text{start}}/d_{\text{end}} = 1.5/0.5$ identified in §\ref{sec:sweep} unchanged to the main results, evaluating it across four architectures (Transformer, Gated Attention, Hope-attention, and Titans) and two scales ($760$M and $1.3$B parameters). Within each pair, total parameter count, training FLOPs, and inference FLOPs are held fixed; only the per-layer MLP intermediate dimension differs. Table~\ref{tab:lm_results} reports WikiText and LAMBADA perplexity along with accuracy on eight commonsense benchmarks.

Across all four architectures and both scales, the tapered model improves average commonsense accuracy over its uniform counterpart, without exception. LAMBADA perplexity improves in all eight comparisons, and WikiText perplexity improves in seven. That the gain holds across this set is more than a robustness check on a single architecture: the motivating evidence for non-uniform layer importance was drawn largely from transformers, but the prescription that follows transfers to softmax attention, gated attention, recurrent self-modifying memory, and neural long-term memory. One plausible reading is that the principle concerns how parameters are allocated across depth in the MLP stack, rather than any property specific to the token-mixing module.

The improvement is also consistent across scales. Rather than diminishing at larger model sizes, every architecture sees gains in average commonsense accuracy at both $760$M and $1.3$B, with perplexity moving in the same direction across configurations. This scale consistency suggests that depth-aware capacity allocation does not saturate with model size, and that tapering remains a free improvement at matched parameters and FLOPs across the operating points studied.

\begin{table*}[t]
\centering
\setlength{\tabcolsep}{4pt}
\renewcommand{\arraystretch}{1.25}
\caption{\textbf{Language modeling and commonsense reasoning results.} We report perplexity on WikiText and LAMBADA and accuracy on eight commonsense benchmarks; Avg.\ denotes the mean accuracy across reasoning tasks. Results are shown for $760$M (50B tokens) and $1.3$B (100B tokens) models, comparing each backbone with and without tapering (shaded rows).}
\label{tab:lm_results}
\vspace{1ex}
\resizebox{\linewidth}{!}{
\begin{tabular}{l cc ccccccccc}
\toprule
& \multicolumn{2}{c}{\textbf{Perplexity}} & \multicolumn{9}{c}{\textbf{Commonsense Reasoning}} \\
\cmidrule(lr){2-3} \cmidrule(lr){4-12}
\textbf{Model} & \textbf{Wiki.} & \textbf{LMB.} & \textbf{LMB.} & \textbf{PIQA} & \textbf{Hella.} & \textbf{Wino.} & \textbf{ARC-e} & \textbf{ARC-c} & \textbf{SIQA} & \textbf{BoolQ} & \textbf{Avg.} \\
& ppl $\downarrow$ & ppl $\downarrow$ & acc $\uparrow$ & acc $\uparrow$ & acc\_n $\uparrow$ & acc $\uparrow$ & acc $\uparrow$ & acc\_n $\uparrow$ & acc $\uparrow$ & acc $\uparrow$ & $\uparrow$ \\
\midrule
\multicolumn{12}{c}{\textit{$760$M params / $50$B tokens}} \\
\midrule
\addlinespace[2pt]
Transformer++ & $21.86$ & $22.29$ & $39.0$ & $68.7$ & $46.3$ & $57.1$ & $66.8$ & $35.3$ & $42.5$ & $62.3$ & $52.25$ \\
\rowcolor{tablerow} \qquad + Tapered & $21.42$ & $21.25$ & $40.1$ & $69.3$ & $47.0$ & $57.3$ & $66.7$ & $35.9$ & $43.0$ & $63.4$ & $\mathbf{52.84}$ \\
\addlinespace[2pt]
Gated Attention & $20.74$ & $21.85$ & $39.7$ & $69.2$ & $46.3$ & $57.9$ & $68.4$ & $35.5$ & $41.8$ & $62.1$ & $52.61$ \\
\rowcolor{tablerow} \qquad + Tapered & $19.98$ & $21.44$ & $40.0$ & $69.3$ & $46.8$ & $57.8$ & $69.1$ & $35.8$ & $41.6$ & $62.6$ & $\mathbf{52.88}$ \\
\addlinespace[2pt]
Hope-attention & $20.62$ & $21.29$ & $40.2$ & $70.1$ & $50.6$ & $56.8$ & $69.9$ & $37.1$ & $41.3$ & $63.5$ & $53.69$ \\
\rowcolor{tablerow} \qquad + Tapered & $20.50$ & $21.07$ & $40.3$ & $70.7$ & $51.0$ & $57.4$ & $69.2$ & $38.1$ & $41.8$ & $63.9$ & $\mathbf{54.05}$ \\
\addlinespace[2pt]
Titans & $21.58$ & $23.09$ & $39.2$ & $67.7$ & $50.0$ & $52.8$ & $68.0$ & $35.6$ & $41.4$ & $63.7$ & $52.30$ \\
\rowcolor{tablerow} \qquad + Tapered & $20.77$ & $22.92$ & $39.9$ & $69.0$ & $51.6$ & $54.5$ & $67.9$ & $36.1$ & $42.5$ & $64.8$ & $\mathbf{53.29}$ \\
\midrule
\multicolumn{12}{c}{\textit{$1.3$B params / $100$B tokens}} \\
\midrule
\addlinespace[2pt]
Transformer++ & $17.39$ & $17.62$ & $45.1$ & $72.8$ & $53.5$ & $59.7$ & $70.6$ & $38.4$ & $44.0$ & $64.3$ & $56.05$ \\
\rowcolor{tablerow} \qquad + Tapered & $17.17$ & $16.93$ & $45.7$ & $72.5$ & $53.4$ & $60.0$ & $70.8$ & $38.3$ & $44.6$ & $65.7$ & $\mathbf{56.38}$ \\
\addlinespace[2pt]
Gated Attention & $16.03$ & $14.26$ & $46.2$ & $73.0$ & $53.9$ & $60.2$ & $71.5$ & $38.7$ & $44.2$ & $64.4$ & $56.51$ \\
\rowcolor{tablerow} \qquad + Tapered & $15.92$ & $14.11$ & $46.5$ & $73.6$ & $53.8$ & $60.4$ & $71.7$ & $38.7$ & $44.5$ & $65.2$ & $\mathbf{56.80}$ \\
\addlinespace[2pt]
Hope-attention & $15.91$ & $15.48$ & $47.0$ & $72.8$ & $54.0$ & $60.3$ & $72.3$ & $38.9$ & $45.1$ & $65.2$ & $56.95$ \\
\rowcolor{tablerow} \qquad + Tapered & $15.94$ & $14.92$ & $47.1$ & $73.0$ & $53.8$ & $60.3$ & $71.9$ & $39.4$ & $45.2$ & $65.7$ & $\mathbf{57.05}$ \\
\addlinespace[2pt]
Titans & $16.05$ & $14.19$ & $46.9$ & $73.1$ & $53.5$ & $59.9$ & $72.4$ & $39.3$ & $43.9$ & $64.8$ & $56.73$ \\
\rowcolor{tablerow} \qquad + Tapered & $15.76$ & $14.04$ & $46.9$ & $73.8$ & $54.2$ & $60.7$ & $72.1$ & $39.6$ & $43.8$ & $65.5$ & $\mathbf{57.08}$ \\
\bottomrule
\end{tabular}
}
\end{table*}

\section{Layer-wise Novelty}
\label{sec:analysis}

The experiments in §\ref{sec:experiments} establish that tapering helps; they do not say why. To probe a mechanism, we measure how much novel information each layer writes into the residual stream of pretrained transformers given uniform capacity. We find that MLP outputs become progressively more aligned with the existing residual as depth increases, reinforcing content already present rather than computing new features. Tapering aligns the architecture with this pattern.

We measure two cosine quantities per layer. Using the notation of §\ref{sec:background} ($z_l = h_l + \mathcal{M}_l(h_l)$, $h_{l+1} = z_l + \mathcal{F}_l(z_l)$),
\begin{align}
\rho_l^{\text{block}} &= \cos\bigl(h_{l+1} - h_l,\, h_l\bigr), \\
\rho_l^{\text{MLP}} &= \cos\bigl(\mathcal{F}_l(z_l),\, h_l\bigr),
\end{align}
averaged over tokens. The block-update quantity captures the layer's full additive contribution; the MLP-only quantity isolates the component being tapered.\footnote{$\mathcal{F}_l(z_l)$ is added to $z_l$ in the residual stream rather than to $h_l$. We use $h_l$ as the reference for both $\rho_l^{\text{block}}$ and $\rho_l^{\text{MLP}}$ to share a single unnormalized residual reference across the two quantities and to avoid the directional distortion that LayerNorm introduces in $z_l$'s normalized input.} $\rho \approx 0$ corresponds to writing content orthogonal to the residual; $\rho > 0$ corresponds to reinforcing a direction the residual already encodes. A rising trend with depth signals diminishing novelty.

\begin{figure}[t]
\centering
\includegraphics[width=\linewidth]{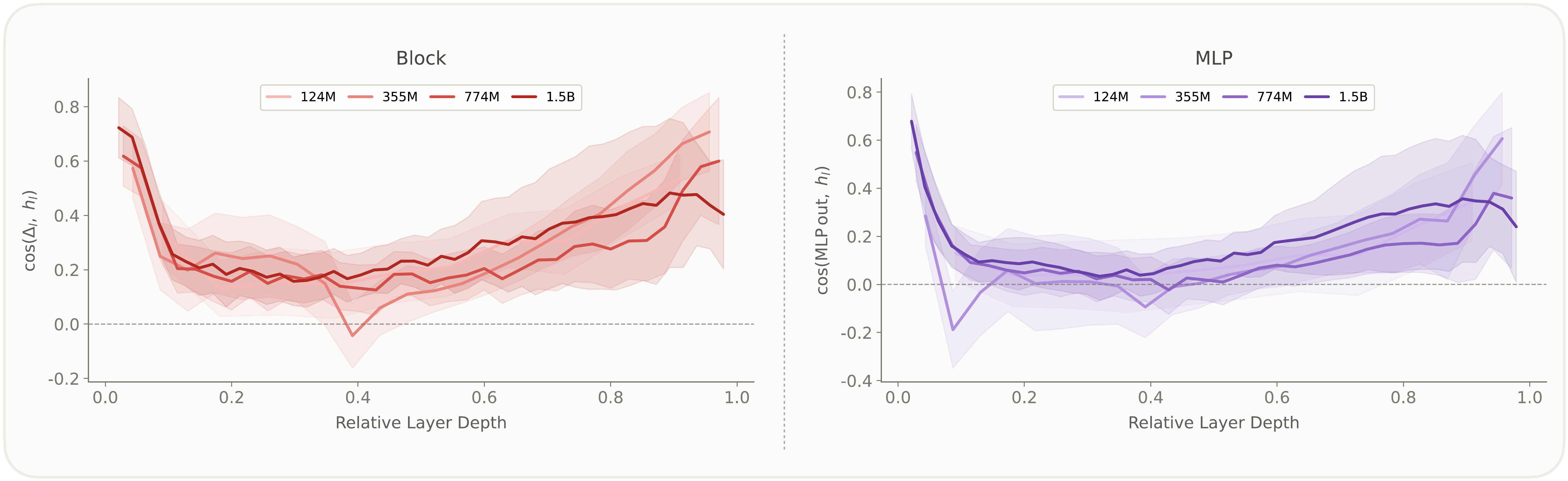}
\caption{\textbf{Layer updates become more aligned with the residual stream at greater depths.} Cosine similarity between each layer's update and the residual stream entering its block, as a function of relative layer depth, across the GPT-2 family ($124$M to $1.5$B; lightest to darkest shade). Left: full block update, $\rho_l^{\text{block}}$. Right: MLP-only, $\rho_l^{\text{MLP}}$. Both rise with depth in every model size, with the MLP signal showing the cleaner monotone trend. Shaded bands show $\pm 1$ standard deviation across tokens; the dashed line marks $\rho = 0$.}
\label{fig:novelty}
\end{figure}

We compute $\rho_l^{\text{block}}$ and $\rho_l^{\text{MLP}}$ on $2048$ tokens of WikiText-2~\citep{merity2016pointer} across the GPT-2~\citep{radford2019language} family, which provides publicly available pretrained checkpoints across a range of scales under a controlled architecture, allowing us to isolate the depth-wise pattern without confounds from training variation. The first and last layers are omitted as boundary cases. The pattern in Figure~\ref{fig:novelty} is consistent across the family: both quantities drop to low values in the early-middle layers and climb through the second half of the network in every model. Pearson correlation with layer index is positive in every measurement and consistently tighter for the MLP than the full block: $\rho_l^{\text{MLP}}$ correlations range from $r = 0.49$ to $r = 0.71$, $\rho_l^{\text{block}}$ from $r = 0.27$ to $r = 0.71$. That both quantities rise is informative on two counts: it establishes that the MLP itself becomes redundant rather than sitting idle while attention does the work, and it shows the depth-wise pattern extends to the layer as a whole rather than being MLP-specific.

The connection to tapering is direct. If later MLPs at uniform capacity produce outputs aligned with the residual rather than orthogonal to it, the additional hidden dimension is not being put to use. Tapering reduces the hidden dimension where this alignment is largest and reallocates the saved parameters to the early layers, where MLP outputs are still orthogonal and added capacity has somewhere to go. This is consistent with the empirical asymmetry of Figure~\ref{fig:motivation}: front-loading capacity helps because early MLPs use it; back-loading hurts because late MLPs cannot. The block-level rise further suggests the principle is not specific to MLPs: any parameter-bearing axis listed in §\ref{sec:tapering} (attention head count, key-value dimension, recurrent state size) is in principle a candidate for tapering, which we leave to future work.

\section{Related Work}
\label{sec:related}

\paragraph{Language model families.}
Across architectural families, today's language models rely on the same per-block design despite differing in their token-mixing modules. Transformers~\citep{vaswani2017attention} and gated-attention variants~\citep{qiu2025gated} use softmax-based attention; linear recurrent and state-space models~\citep{schlag2021linear, tiezzi2024resurgence, sun2023retentive, peng2023rwkv, peng2024eagle, peng2025rwkv7, smith2023simplified, hasani2022liquid, hasani2021liquid, dao2024transformers, beck2024xlstm, orvieto2023resurrecting, de2024griffin, liu2024longhorn, siems2025deltaproduct, ren2025samba, merrill2026olmo} replace it with data-dependent recurrences for sub-quadratic scaling; and memory-based architectures~\citep{sun2024learning, behrouz2024titans, behrouz2025s, behrouz2025nested, behrouz2025atlas, wang2025test, hu2026improving, li2026tnt} augment it with learnable memory modules that adapt at test time. Across all of these, each block pairs a token-mixing module with an MLP, with parameters allocated uniformly across depth. Tapering targets this shared default, which makes the question of how to allocate MLP capacity across depth applicable to every member of the family.

\paragraph{Non-uniform allocations across depth.}
A growing line of work varies architectural dimensions across depth, though along axes other than static MLP width: pooling sequence length~\citep{dai2020funnel}, dynamically routing compute to tokens~\citep{raposo2024mixture, bae2026mixture}, dropping entire layers during training~\citep{fan2019reducing}, and nesting variable-size blocks within layers for elastic inference~\citep{kudugunta2024matformer}. These directions are orthogonal to ours: rather than routing tokens, sharing layers, or varying dimensions within a single block, we ask whether the MLP intermediate dimension should vary monotonically across layers as a fixed architectural choice.

A closer cluster redistributes parameters across layers. Block-wise scaling~\citep{mehta2020delight} varies per-layer FFN multipliers and head counts jointly across depth. \citet{baroian2025crown} ablate several layer-wise scaling shapes at fixed budget but stop short of identifying a clear winner. \citet{ikeda2025layerwise} take a more aggressive approach, deactivating FFNs in some layers entirely and concentrating capacity in a contiguous band; their headline finding favors middle placement, though their own per-scale results vary, with early-heavy configurations rising to the top in their deepest setting. These works agree that uniform allocation is suboptimal but reach inconsistent conclusions about the right shape, with the discrepancies likely reflecting differences in setup: uncontrolled comparisons against unequal-cost baselines, joint scaling of multiple architectural axes, or binary allocations that change effective depth. Tapering keeps every layer active and isolates a single axis of variation, MLP width, against an equal-cost uniform baseline.

\paragraph{Layer importance.}
A separate line of work documents that the contribution of layers to the final output is not uniform across depth. Early-exit and layer-skipping methods show that the residual stream typically reaches its final prediction before the last layer~\citep{elbayad2020depth, schuster2022confident, elhoushi2024layerskip, belrose2023eliciting}, and structured redundancy analyses find that many layers, particularly later ones, can be removed with little performance loss~\citep{men2025shortgpt, gromov2024unreasonable, lad2024remarkable}. Interpretability work points to a related shift in the nature of computation across depth, with FFNs functioning as key-value memories whose roles change from shallow to semantic patterns~\citep{geva2021transformer}, attention heads showing comparable redundancy~\citep{voita2019analyzing}, and intermediate layers carrying richer representations than final layers~\citep{skean2025layer}. Activation-steering studies provide complementary intervention-based evidence, finding that behavioral steerability is strongest in intermediate layers and therefore varies substantially with depth~\citep{bayat2025steering}. Post-training methods exploit this non-uniformity, applying selective rank reduction~\citep{sharma2023truth} or structured pruning~\citep{ashkboos2024slicegpt} to MLP weights, but require additional pruning and fine-tuning stages. Tapering is orthogonal to capacity-allocation methods that vary across tokens, such as Mixture-of-Experts~\citep{fedus2022switch}, and could in principle be combined with them. We complement these analyses with a direct measurement showing that MLP outputs become progressively more aligned with the residual stream at greater depths (§\ref{sec:analysis}), providing a mechanistic rationale for the tapering direction.

\section{Limitations}
\label{sec:limitations}

Our schedule and width-ratio sweep is conducted only on the $440$M Transformer. We then transfer the selected cosine schedule with $d_{\text{start}}/d_{\text{end}} = 1.5/0.5$ unchanged to the larger models and alternative architectures. This design provides a deliberately strict test of transferability: the improvements in the main experiments show that the selected configuration is a robust default, but they do not establish that it is optimal for every model.

The preferred tapering profile may depend on properties such as model depth, hidden dimension, the fraction of parameters allocated to the MLP, the token-mixing architecture, or the training budget. In particular, the best schedule or endpoint ratio may shift as models scale, potentially yielding gains beyond those reported here. A broader study that sweeps tapering configurations across model sizes and architectural families would help characterize these interactions and determine whether a single transferable schedule is sufficient or architecture- and scale-specific configurations are preferable.

\section{Discussion and Conclusion}
\label{sec:conclusion}

We introduced \textit{Tapered Language Models} (TLMs), an architectural principle in which a parameter-bearing component is monotonically tapered across depth under a fixed total budget. Instantiated through a smooth cosine taper over MLP intermediate width, tapering improves perplexity and downstream benchmark performance relative to uniform-width baselines across three model scales and four architectural families spanning softmax attention, gated attention, recurrent self-modifying memory, and neural long-term memory, and provides mechanistic evidence that MLP outputs become progressively more aligned with the residual stream at greater depths, directly motivating the tapering direction. These gains are achieved without increasing parameter count or training compute. Across our motivating experiments, schedule ablations, and width sweeps, a consistent picture emerges: model capacity is most valuable in earlier layers, and smoothly decaying allocations provide an effective and robust way to distribute it.

We view these results as an initial exploration of a broader architectural principle. Beyond MLP width, many other layer-wise dimensions that determine parameter allocation---such as attention head count, key-value dimension, recurrent state size, memory slot count, and expert count in mixture-of-experts models---are natural candidates for tapering. Whether similar gains extend across these axes remains an open empirical question.

More broadly, the assumptions targeted by tapering extend beyond language modeling. Vision transformers~\citep{dosovitskiy2020image}, diffusion transformers~\citep{peebles2023scalable}, and multimodal models~\citep{radford2021learning, liu2023visual} all inherit the same default pattern of approximately uniform capacity across depth. Our results suggest that this convention may be unnecessarily restrictive. We therefore see depth-aware capacity allocation as a simple, low-cost architectural design lever with potential applications across the broader foundation model landscape.

\bibliography{paper}

\clearpage
\appendix
\section{Long-Context Retrieval on Needle-in-a-Haystack}
\label{app:niah}

We evaluate long-context retrieval on Needle-in-a-Haystack (NIAH) to verify that redistributing MLP capacity across depth does not degrade long-context behavior. We use three single-needle variants of increasing difficulty (S-NIAH-1 retrieves a passkey, S-NIAH-2 a numerical value, and S-NIAH-3 a UUID) and a multi-query variant (MQ-NIAH), each evaluated at three context lengths ($4$K, $8$K, and $16$K). Results are in Table~\ref{tab:niah}. Tapered models match or improve over their uniform counterparts across the table, with gains concentrating in the harder cells where absolute scores are lowest.

\begin{table*}[t]
\centering
\setlength{\tabcolsep}{4pt}
\renewcommand{\arraystretch}{1.25}
\caption{\textbf{Long-context retrieval results.} Needle-in-a-Haystack experiments with three single-needle variants of increasing difficulty (S-NIAH-1 retrieves a passkey, S-NIAH-2 a numerical value, and S-NIAH-3 a UUID) and a multi-query variant (MQ-NIAH), evaluated at three context lengths. All entries are retrieval accuracy; higher is better. Tapered rows are shaded.}
\label{tab:niah}
\resizebox{\linewidth}{!}{
\begin{tabular}{l ccc ccc ccc ccc}
\toprule
& \multicolumn{3}{c}{S-NIAH-1} & \multicolumn{3}{c}{S-NIAH-2} & \multicolumn{3}{c}{S-NIAH-3} & \multicolumn{3}{c}{MQ-NIAH} \\
& \multicolumn{3}{c}{(passkey)} & \multicolumn{3}{c}{(number)} & \multicolumn{3}{c}{(UUID)} & \multicolumn{3}{c}{(multi-query)} \\
\cmidrule(lr){2-4} \cmidrule(lr){5-7} \cmidrule(lr){8-10} \cmidrule(lr){11-13}
Model & 4K & 8K & 16K & 4K & 8K & 16K & 4K & 8K & 16K & 4K & 8K & 16K \\
\midrule
\addlinespace[2pt]
Transformer++ & $96.4$ & $84.6$ & $82.4$ & $100$ & $99.4$ & $95.8$ & $80.8$ & $72.4$ & $47.2$ & $56.8$ & $45.4$ & $27.2$ \\
\rowcolor{tablerow} \quad + Tapered & $96.4$ & $84.8$ & $82.8$ & $100$ & $99.2$ & $96.4$ & $81.4$ & $72.4$ & $47.8$ & $57.4$ & $45.4$ & $27.6$ \\
\addlinespace[2pt]
Gated Attention & $97.2$ & $86.4$ & $84.0$ & $100$ & $99.8$ & $95.8$ & $81.2$ & $72.2$ & $47.4$ & $57.2$ & $46.0$ & $27.2$ \\
\rowcolor{tablerow} \quad + Tapered & $97.2$ & $86.8$ & $84.0$ & $100$ & $99.6$ & $96.2$ & $81.2$ & $72.4$ & $47.8$ & $57.8$ & $46.2$ & $27.2$ \\
\addlinespace[2pt]
Hope-attention & $100$ & $100$ & $100$ & $99.8$ & $99.2$ & $96.4$ & $83.6$ & $72.8$ & $49.8$ & $61.4$ & $47.4$ & $30.8$ \\
\rowcolor{tablerow} \quad + Tapered & $100$ & $100$ & $100$ & $100$ & $99.8$ & $96.4$ & $83.2$ & $73.2$ & $50.0$ & $61.6$ & $47.8$ & $31.2$ \\
\addlinespace[2pt]
Titans & $100$ & $100$ & $100$ & $99.2$ & $83.8$ & $67.8$ & $74.4$ & $38.8$ & $18.2$ & $21.2$ & $19.2$ & $11.8$ \\
\rowcolor{tablerow} \quad + Tapered & $100$ & $100$ & $100$ & $99.2$ & $84.2$ & $68.4$ & $74.2$ & $39.2$ & $20.4$ & $21.8$ & $21.0$ & $12.4$ \\
\bottomrule
\end{tabular}
}
\end{table*}

\end{document}